\documentclass[12pt,a4]{article}
\usepackage[dvips]{color,graphicx}
\usepackage{bm,psfrag,afterpage,color,amsmath, amssymb,latexsym,amsthm}

\renewcommand{\arraystretch}{0.8}

\renewcommand{\arraystretch}{1.0} 

\newcommand{\argmax}{\mathop{\rm arg~max}\limits}

\def\Cov{{\rm Cov}}
\def\Corr{{\rm Corr}}

\makeatletter
\def\eqnarray{\stepcounter {equation}\let \@currentlabel =\theequation
\global \@eqnswtrue
\global \@eqcnt \z@ \tabskip \@centering \let \\=\@eqncr
$$\halign to \displaywidth \bgroup \@eqnsel \hskip \@centering
$\displaystyle \tabskip \z@ {##}$&\global \@eqcnt \@ne \hfil
${\mbox{}##\mbox{}}$\hfil &\global \@eqcnt \tw@
$\displaystyle \tabskip \z@ {##}$\hfil \tabskip \@centering
&\llap {##}\tabskip \z@ \cr}
\makeatother

\renewcommand{\arraystretch}{0.9}

\begin{document}

{\baselineskip = 8mm 

\begin{center}
\textbf{\LARGE Sparse principal component regression with adaptive loading} 
\end{center}
\begin{center}
{\large Shuichi Kawano$^{1,4}$, \ Hironori Fujisawa$^{2,4}$, \\
Toyoyuki Takada$^{3,4}$ \ and \ Toshihiko Shiroishi$^{3,4}$}
\end{center}

\begin{center}
\begin{minipage}{14cm}
{
\begin{center}
{\it {\footnotesize 

\vspace{1.2mm}


$^1$ Graduate School of Information Systems,  University of Electro-Communications, \\
1-5-1, Chofugaoka, Chofu-shi, Tokyo 182-8585, Japan. \\

\vspace{1.2mm}

$^2$ The Institute of Statistical Mathematics, \\
10-3 Midori-cho, Tachikawa, Tokyo 190-8562, Japan. \\

\vspace{1.2mm}

$^3$ Mammalian Genetics Laboratory, 
National Institute of Genetics, \\ Mishima, Shizuoka 411-8540, Japan.\\

\vspace{1.2mm}

$^4$ Transdisciplinary Research Integration Center, \\
Research Organization of Information and Systems, Minato-ku, Tokyo 105-0001, Japan. \\
}}

\vspace{2mm}

skawano@ai.is.uec.ac.jp \hspace{5mm} fujisawa@ism.ac.jp \\
ttakada@nig.ac.jp \hspace{5mm} tshirois@nig.ac.jp

\end{center}

\vspace{1mm} 

{\small {\bf Abstract:} \
Principal component regression (PCR) is a two-stage procedure that selects some principal components and then constructs a regression model regarding them as new explanatory variables. 
Note that the principal components are obtained from only explanatory variables and not considered with the response variable. 
To address this problem, we propose the sparse principal component regression (SPCR) that is a one-stage procedure for PCR. 
SPCR enables us to adaptively obtain sparse principal component loadings that are related to the response variable and select the number of principal components  simultaneously. 
SPCR can be obtained by the convex optimization problem for each parameter with the coordinate descent algorithm. 
Monte Carlo simulations and real data analyses are performed to illustrate the effectiveness of SPCR. 
 }

\vspace{3mm}



}
\end{minipage}
\end{center}

\baselineskip = 8mm


\section{Introduction}

Principal component analysis (PCA) (Jolliffe, 2002) is a fundamental statistical tool for dimensionality reduction, data processing, and  visualization of multivariate data, with various applications in biology, engineering, and social science. 
In regression analysis, it can be useful to replace many original explanatory variables with a few principal components, which is called the principal component regression (PCR) (Massy, 1965; Jolliffe, 1982). 
PCR is widely used in various fields of research and many extensions of PCR have been proposed (see, e.g., Hartnett {\it et al.}, 1998; Rosital \textit{et al.}, 2001; Reiss and Ogden, 2007; Wang and Abbott, 2008). 
Whereas PCR is a useful tool for analyzing multivariate data, this method may not have enough prediction accuracy if the response variable depends on the principal components with small eigenvalues. 
The problem arises from the two-stage procedure for PCR; a few principal components are selected with large eigenvalues, but without any relation to response variable, and then the regression model is constructed using them as new explanatory variables.

In this paper, we deal with PCA and regression analysis simultaneously, and propose a one-stage procedure for PCR to address this problem. 
The procedure combines two loss functions; one is the ordinary regression analysis loss and the other is PCA loss  with some devices proposed by Zou \textit{et al.} (2006). 
In addition, in order to easily interpret estimated principal component loadings and select the number of principal components automatically, we impose the $L_1$ type regularization on the parameters. 
This one-stage procedure is called the sparse principal component regression (SPCR) in this paper. 
SPCR gives sparse principal component loadings that are related to the response variable and selects the number of principal components simultaneously. 
We also establish a monotonically decreasing estimation procedure for the loss function using the coordinate descent algorithm (Friedman {\it et al.}, 2010), because SPCR can be obtained via the convex optimization problem for each of parameters. 

The partial least squares regression (PLS) (Wold, 1975; Frank and Friedman, 1993) is a dimension reduction technique, which incorporates information between  the explanatory variables and the response variable. 
Recently, Chun and Kele\c{s} (2010) have proposed the sparse partial least squares regression (SPLS) that imposes sparsity in the dimension reduction step of PLS, and then constructed a regression model regarding some SPLS components as new explanatory variables, although it is a two-stage procedure.  
Besides PLS and SPLS, several methods have been proposed for performing dimension reduction and regression analysis simultaneously. 
Bair {\it et al.} (2006) proposed the supervised principal component analysis, which is regression analysis in which the explanatory variables are related to the response variable with respect to correlation. 
Yu {\it et al.} (2006) presented the supervised probabilistic principal component analysis from the Bayesian viewpoint. 
By imposing the $L_1$ type regularization into the objective function, Allen {\it et al.} (2013) and Chen and Huang (2012) introduced the regularized partial least squares and the sparse reduced-rank regression, respectively. 
However, none of them integrated the two loss functions for ordinary regression analysis and PCA along with the $L_1$ type regularization.

This paper is organized as follows. 
In Section 2, we review PCA and the sparse principal component analysis (SPCA) by Zou \textit{et al.} (2006). 
We propose SPCR and discuss alternative methods to SPCR in Section 3. 
Section 4 provides an efficient algorithm for SPCR and a method for selecting tuning parameters in SPCR. 
Monte Carlo simulations and real data analyses are provided in Section 5. 
Concluding remarks are given in Section 6. 
The R language software package {\tt spcr}, which implements SPCR, is available on the Comprehensive R Archive Network ({\it http://cran.r-project.org}). 
Supplementary materials can be found in {\it https://sites.google.com/site/shuichikawanoen/research/suppl\_spcr.pdf}.

\section{Preliminaries}

\subsection{Principal component analysis}
Let $X=({\bm x}_1, \ldots, {\bm x}_n)^T$ be an $n \times p$ data matrix, where $n$ and $p$ denote the sample size and the number of variables, respectively. 
Without loss of generality, we assume that the column means of the matrix $X$ are all zero.

PCA is usually implemented by using the singular value decomposition (SVD) of $X$. 
When the SVD of $X$ is represented by 
\begin{eqnarray*}
X=UDV^T,
\end{eqnarray*} 
the principal components are $Z = UD$ and the corresponding loadings of the principal components are the columns of $V$. 
Here, $U$ is an $n \times n$ orthogonal matrix, $V=({\bm v}_1,\ldots,{\bm v}_p)$ is a $p \times p$ orthogonal matrix, and $D$ is an $n \times p$ matrix given by
\begin{eqnarray*}
D = \left( 
\begin{array}{cc}
D^* & O_{q, p-q}  \\
O_{n-q, q} & O_{n-q, p-q}  \\
\end{array} 
\right),
\end{eqnarray*}
where $q={\rm rank} (X)$, $D^* = {\rm diag} (d_1,\ldots,d_q) \ (d_1 \geq \ldots \geq d_q>0)$, and $O_{i,j}$ is the $i \times j$ matrix with all zero elements. 
Note that the vectors $V^T {\bm x}_1, \ldots, V^T {\bm x}_n$ are also the principal components, since $XV = Z$.

The loading matrix can be obtained by solving the following least squares problem (see, e.g., Hastie {\it et al.}, 2009); 
\begin{eqnarray}
&& \min_{B} \sum_{i=1}^n || {\bm x}_i - B B^T {\bm x}_i ||^2  \label{pca} \\
&& subject \ to \ \ \ B^T B = I_{k}, \nonumber
\end{eqnarray}
where $B=({\bm \beta}_1,\ldots,{\bm \beta}_k)$ is a $p \times k$ loading matrix, $k$ denotes the number of principal components, and $I_k$ is the $k \times k$ identity matrix. 
The solution is given by
\begin{eqnarray*}
\hat{B} = V_k Q^T,
\end{eqnarray*}
where $V_k = ({\bm v}_1,\ldots, {\bm v}_k)$ and $Q$ is a $k \times k$ arbitrary orthogonal matrix. 
\subsection{Sparse principal component analysis}
Zou {\it et al.} (2006) proposed an alternative least squares problem given by
\begin{eqnarray}
&& \min_{A, B} \sum_{i=1}^n || {\bm x}_i - A B^T {\bm x}_i ||^2 + \lambda \sum_{j=1}^k || {\bm \beta}_j ||^2  \label{pca2} \\
&& subject \ to \ \ \ A^T A = I_{k}, \nonumber
\end{eqnarray}
where $A=({\bm \alpha}_1,\ldots,{\bm \alpha}_k)$ is a $p \times k$ matrix and $\lambda \ (>0)$ is a regularization parameter.  
The minimizer of $B$ is given by
\begin{eqnarray}
\hat{B} = V_k C Q^T,
\label{solution1}
\end{eqnarray}
where $C={\rm diag} (c_1,\ldots,c_k)$, $c_i \ (i=1,\ldots,k)$ is a positive constant, and $Q$ is an arbitrary orthogonal matrix. 
The case $\lambda=0$ yields the same solution as (\ref{pca}). 
Formula (\ref{pca2}) is a quadratic programming problem with respect to each parameter matrix $A$ and $B$, but Formula (\ref{pca}) is not. 

In addition, Zou {\it et al.} (2006) proposed to add a sparse regularization term for $B$ to easily interpret the estimate $\hat{B}$, which is  called SPCA; 
\begin{eqnarray}
&& \min_{A, B} \sum_{i=1}^n || {\bm x}_i - A B^T {\bm x}_i ||^2 + \lambda \sum_{j=1}^k || {\bm \beta}_j ||^2 + \sum_{j=1}^k \lambda_{1,j} || {\bm \beta}_j ||_1 \label{spca} \\
&& subject \ to \ \ \ A^T A = I_{k}, \nonumber
\end{eqnarray}
where $\lambda_{1,j}$'s \ $(j=1,\ldots,k)$ are regularization parameters with positive value and $|| \cdot ||_1$ is the $L_1$ norm of $\bm \beta$. 
Note that the minimization problem (\ref{spca}) is also the quadratic programming problem with respect to each parameter matrix $A$ and $B$. 
After simple calculation, the problem (\ref{spca}) becomes
\begin{eqnarray*}
&& \min_{A, B} \sum_{j=1}^k  \left\{ || X {\bm \alpha}_j - X {\bm \beta}_j  ||^2 + \lambda || {\bm \beta}_j ||^2 +  \lambda_{1,j} || {\bm \beta}_j ||_1 \right\} \\
&& subject \ to \ \ \ A^T A = I_{k}. 
\end{eqnarray*}
This optimization problem is analogous to the elastic net problem in Zou and Hastie (2005), and hence Zou {\it et al.} (2006) proposed an alternating algorithm to estimate $A$ and $B$ iteratively. 
In particular, the LARS algorithm (Efron {\it et al.}, 2004) is employed to obtain the estimate of $B$ numerically.

Another approach to obtain a sparse loading matrix is SCoTLASS (Jolliffe {\it et al}., 2003). 
However, Zou {\it et al.} (2006) pointed out that the loadings obtained by SCoTLASS are not sparse enough. 
Also, Lee {\it et al.} (2010) and Lee and Huang (2013) developed SPCA for binary data.

\section{Sparse principal component regression}

\subsection{Sparse principal component regression with adaptive loading}
\label{Sparse principal component regression}
Suppose that we have data for response variables $y_1,\ldots,y_n$ in addition to data ${\bm x}_1,\ldots, {\bm x}_n$. 
We consider regression analysis in the situation that the response variable is explained by variables aggregated by PCA of $X=({\bm x}_1,\ldots, {\bm x}_n)^T$. 
A naive approach is to construct a regression model with a few principal components corresponding to large eigenvalues, which are previously constructed. 
This approach is called PCR. 
In general, principal components are irrelevant with the response variables. 
Therefore, PCR might fail to predict the response if the response is associated with principal components corresponding to small eigenvalues.

To overcome this drawback, we propose SPCR using the principal components $B^T {\bm x}$ as follows: 
\begin{eqnarray}
&& \min_{A, B, \gamma_0, {\bm \gamma}} \Big\{(1-w) \sum_{i=1}^n \left(y_i - \gamma_0 - {\bm \gamma}^T B^T {\bm x}_i \right)^2 + w \sum_{i=1}^n || {\bm x}_i - A B^T {\bm x}_i ||^2   \nonumber \\
&&  \hspace{30mm} + \lambda_{\beta} (1-\zeta) \sum_{j=1}^k || {\bm \beta}_j ||_1 + \lambda_{\beta} \zeta \sum_{j=1}^k || {\bm \beta}_j ||^2 + \lambda_{\gamma} || {\bm \gamma} ||_1\Big\}  \label{spcr} \\
&& subject \ to \ \ \ A^T A = I_{k}, \nonumber
\end{eqnarray}
where $\gamma_0$ is an intercept, ${\bm \gamma} = (\gamma_1,\ldots,\gamma_k)^T$ is a coefficient vector, $ \lambda_{\beta}$ and $\lambda_\gamma$ are regularization parameters with positive value, and $w$ and $\zeta$ are tuning parameters whose values are between zero and one.

The first term in Formula (\ref{spcr}) means the least squares loss between the response and the principal components $B^T {\bm x}$.  
The second term induces PCA loss of data $X$. 
The tuning parameter $w$ controls the trade-off between the first and second terms, and then the value of $w$ can be determined by users for any purpose. 
For example, a smaller value for $w$ is used when we aim to obtain better prediction accuracies, while a larger value for $w$ is used when we aim to obtain the exact formulation of the principal component loadings. 
The third and fifth terms encourage sparsity on $B$ and ${\bm \gamma}$, respectively. 
The sparsity on $B$ enables us to easily interpret the loadings of the principal components. 
Meanwhile, the sparsity on $\bm \gamma$ induces automatic selection of the number of principal components. 
The tuning parameter $\zeta$ controls the trade-off between the $L_1$ and $L_2$ norms for the parameter $B$, which was introduced in Zou and Hastie (2005). 
For detailed roles of this parameter and the $L_2$ norm, see Zou and Hastie (2005).

We see that (\ref{spcr}) is a quadratic programming problem with respect to each parameter, because the problem only combines a regression loss with PCA loss. 
The optimization problem  appears to be simple. 
However, it is not easy to numerically obtain the estimates of the parameters if we do not introduce the $L_1$ regularization terms for $B$ and ${\bm \gamma}$, because there exists an identification problem for $B$ and ${\bm \gamma}$. 
For an arbitrary orthogonal matrix $P$, we have 
\begin{eqnarray*}
{\bm \gamma}^T B^T = {\bm \gamma}^T P^T P B^T = {\bm \gamma}^{\dag T} B^{\dag T},
\end{eqnarray*}
where ${\bm \gamma}^{\dag} = P {\bm \gamma}$ and $B^{\dag} = B P^T$. 
This causes non-unique estimators for $B$ and ${\bm \gamma}$. 
However, we incorporate the $L_1$-penalties on (\ref{spcr}) and then we can expect to obtain the minimizer, because the parameter exists on a hypersphere due to orthogonal invariance in (\ref{solution1}) and the $L_1$-penalty implies a hypersquare region. 
For more details, see, e.g.,  Tibshirani (1996),  Jennrich (2006), Choi {\it et al.} (2011), and Hirose and Yamamoto (2014). 
The $L_1$-penalties on $B$ and ${\bm \gamma}$ play two types of roles on sparsity and identification problem. 


\subsection{Adaptive sparse principal component regression}
In the numerical study in Sect. \ref{NumericalStudy}, we observe that SPCR does not produce enough sparse solution for the loading matrix $B$. 
We, therefore, assign different weights to different parameters in the loading matrix $B$. 
This idea was adopted in the adaptive lasso (Zou, 2006). 
Let us consider the weighted sparse principal component regression, given by
\begin{eqnarray*}
&& \min_{A, B, \gamma_0, {\bm \gamma}} \Big\{(1-w) \sum_{i=1}^n (y_i - \gamma_0 - {\bm \gamma}^T B^T {\bm x}_i)^2 + w \sum_{i=1}^n || {\bm x}_i - A B^T {\bm x}_i ||^2    \\
&&  \hspace{30mm} + \lambda_{\beta} (1-\zeta) \sum_{j=1}^k  \sum_{l=1}^p \omega_{lj} |\beta_{lj}| + \lambda_{\beta} \zeta \sum_{j=1}^k || {\bm \beta}_j ||^2 + \lambda_{\gamma} || {\bm \gamma} ||_1\Big\} \label{aspcr} \\
&& subject \ to \ \ \ A^T A = I_{k}, 
\end{eqnarray*}
where $\omega_{lj} \ (>0)$ is an incorporated weight for the parameter $\beta_{lj}$. 
We call this procedure the adaptive sparse principal component regression (aSPCR). 
In this paper, we define the weight as $\omega_{lj} = 1/ |\hat{\beta}_{lj} ({\rm SPCR}) |$, where $\hat{\beta}_{lj} ({\rm SPCR)}$ is an estimate of the parameter ${\beta}_{lj}$ obtained from SPCR. 
In the adaptive lasso, the weight is constructed using the least squares estimators, but it is not applicable due to the identification problem, as described in Sect. \ref{Sparse principal component regression}.

Since aSPCR is a quadratic programming problem with respect to each parameter, we can estimate the parameters according to an efficient estimation algorithm for SPCR. 
In addition, aSPCR enjoys properties similar to SPCR as described in Sect. \ref{Sparse principal component regression}.

\subsection{Related work}
\label{Related work}
PLS  (see, e.g., Wold, 1975; Frank and Friedman, 1993)  seeks directions that relate $X$ to ${\bm y}$ and capture the most variable directions in the $X$-space, which is, in general,  formulated by
\begin{eqnarray}
&& {\bm w}_k = \argmax_{\bm w} \left[ {\rm Corr}^2 ({\bm y}, X {\bm w}) {\rm Var} (X {\bm w}) \right] \label{PLS} \\
&& subject \ to \ \ \ {\bm w}^T {\bm w} = 1, \quad {\bm w}^T \Sigma_{XX} {\bm w}_j = 0, \ \ j=1,\ldots,k-1 \nonumber
\end{eqnarray}
for $k=1,\ldots,p$, where ${\bm y}=(y_1,\ldots,y_n)^T$ and $\Sigma_{XX}$ is the covariance matrix of $X$. 
The solutions in the problem (\ref{PLS}) are derived from NIPALS (Wold, 1975) or SIMPLS (de Jong, 1993).

To incorporate sparsity into PLS, SPLS was introduced by Chun and Kele\c{s} (2010). 
The first SPLS direction vector ${\bm c}$ is obtained by
\begin{eqnarray}
&& \min_{\bm w, \bm c} \left\{ - \kappa {\bm w}^T M {\bm w} + (1-\kappa) ({\bm c} - {\bm w})^T M ({\bm c} - {\bm w}) + \lambda_{1,{\rm SPLS}} ||{\bm c}||_1 +  \lambda_{2,{\rm SPLS}} ||{\bm c}||^2 \right\} \label{spls1} \\
&& subject \ to \ \ \ {\bm w}^T {\bm w} = 1, \nonumber
\end{eqnarray}
where $M=X^T {\bm y} {\bm y}^T X$, and $\kappa, \lambda_{1,{\rm SPLS}}, \lambda_{2,{\rm SPLS}}$ are tuning parameters with positive value. 
Note that the problem (\ref{spls1}) becomes the original maximum eigenvalue problem of PLS when $\kappa=1$, $\lambda_{1,{\rm SPLS}}=0$, and $\lambda_{2,{\rm SPLS}}=0$. 
This SPLS problem is solved by alternately estimating the parameters ${\bm w}$ and ${\bm c}$. 
The idea is similar to that used in SPCA. 
Chun and Kele\c{s} (2010) furthermore introduced the SPLS-NIPALS and SPLS-SIMPLS algorithm for deriving the rest of the direction vectors, and then predicted the response variable by a linear model with SPLS loading vectors as new explanatory variables; it is a two-stage procedure. 

To emphasize a difference between our proposed method and the related work described above, we consider an example as follows. 
Suppose that
\[
y = a_1x_1+a_2x_2+\varepsilon, \qquad x_j \sim N(0,\tau_j^2), \quad \varepsilon \sim N(0,\sigma^2).
\]
This model has another expression in the form
\[
y=a_1^* z_1 + a_2^* z_2 + \varepsilon, \qquad z_j \sim N(0,1), \quad a_j^*=a_j \tau_j.
\]
The covariance structures are given by
\[
\Cov(y,x_j)=a_j \tau_j^2, \qquad \Cov(y,z_j)=a_j^* = a_j \tau_j.
\]
Let us select the explanatory variable that maximizes the covariance:
\begin{eqnarray*}
\max_{x} \Cov(y,x) \qquad {\rm or} \qquad \max_{z} \Cov(y,z) = \max_{z} \Corr(y,z).
\end{eqnarray*}
Consider the case $(a_1,a_2,\tau_1,\tau_2)=(8,1,1,3)$. 
It follows that $a_1^*=8, \ a_2^*=3, \ a_1 \tau_1^2=8$ and $a_2 \tau_2^2=9$. 
In this case, it is clear that the first variable $(x_1,z_1)$ has a larger effect in $y$ than the second variable $(x_2,z_2)$. 
Remember that PLS and SPLS are based on the maximization of covariance, so that they will firstly select the variable $z_1$ on the second maximization, whereas they will firstly select the variable $x_2$ on the first maximization. 
Therefore, on the first maximization, PLS and SPLS fail to select the explanatory variable largely associated with the response. 
Meanwhile, SPCR will select the first variable $(x_1, z_1)$ on both maximizations, because the prediction error remains unchanged after normalization. 


\section{Implementation}
\subsection{Computational algorithm}
For estimating the parameter $A$, we utilize the same algorithm given by Zou \textit{et al.} (2006). 
The parameters $B$ and ${\bm \gamma}$ are estimated by the coordinate descent algorithm (Friedman {\it et al.}, 2010), because the optimization problems include the $L_1$ regularization terms, respectively.

The optimization problem in aSPCR is rewritten as follows:
\begin{eqnarray*}
&& \min_{A, B, \gamma_0, {\bm \gamma}} \Bigg[ (1-w) \sum_{i=1}^n \left\{ y_i - \gamma_0 -  \sum_{j=1}^k \gamma_j \left( \sum_{l=1}^p \beta_{lj} x_{il} \right) \right\}^2 + w \sum_{j=1}^k \sum_{i=1}^n \left( y^*_{ji} - \sum_{l=1}^p \beta_{lj} x_{il} \right)^2    \\ 
&& \hspace{30mm}  + \lambda_{\beta} (1-\zeta) \sum_{j=1}^k \sum_{l=1}^p \omega_{lj} |\beta_{lj}| + \lambda_{\beta} \zeta \sum_{j=1}^k \sum_{l=1}^p \beta_{lj}^2 + \lambda_{\gamma} \sum_{j=1}^k |\gamma_j| \Bigg]  \\
&& subject \ to \ \ \ A^T A = I_{k}, 
\end{eqnarray*}
where $y^*_{ji}$ is the $i$-th element of the vector $X {\bm \alpha}_j$. 
SPCR is a special case of aSPCR with $\omega_{lj}=1$. 
The detailed algorithm is given as follows.

\begin{description}
\item[ $\beta_{lj}$ given $\gamma_0$, $\gamma_j$ and $A$:] 
The coordinate-wise update for $\beta_{lj}$ has the form:
\begin{eqnarray}
&& \hat{\beta}_{l^{\prime} j^{\prime}} \leftarrow \frac{S\left( \sum_{i=1}^n x_{i l^{\prime}} \left\{ (1-w) Y_i \gamma_{j^{\prime}} + Y_{j^{\prime} i }^* w\right\}, \frac{\lambda_{\beta} \omega_{l^{\prime} j^{\prime}} (1-\zeta)}{2} \right)}{ \left\{ (1-w)\gamma_{j^{\prime}}^2 + w\right\} \sum_{i=1}^n x^2_{il^{\prime}} + \lambda_{\beta} \zeta}, \label{UpdateBeta} \\
&& \hspace{80mm} (l^{\prime} = 1,\ldots,p; \ j^{\prime} = 1,\ldots,k), \nonumber
\end{eqnarray}
where
\begin{eqnarray*}
Y_i &=&  y_i - \gamma_0 - \sum_{j=1}^k \sum_{l \neq l^{\prime}} \gamma_j \beta_{lj} x_{il} - \sum_{j \neq j^{\prime}} \gamma_j \beta_{l^{\prime} j} x_{i l^{\prime}}, \\
Y^*_{j^{\prime} i} &=& y_{j^{\prime} i}^* - \sum_{l \neq l^{\prime}} \beta_{l j^{\prime}} x_{il},
\end{eqnarray*}
and $S(z, \eta)$ is the soft-threshholding operator with value
\begin{eqnarray*}
{\rm sign} (z) (|z| - \eta)_+   = \left\{ \begin{array}{ll}
z - \eta & (z > 0 \ {\rm and} \ \eta < |z|) \\
z + \eta & (z < 0 \ {\rm and} \ \eta < |z|) \\
0 & (\eta \geq |z|). \\
\end{array} \right.
\end{eqnarray*}

\item[ $\gamma_j$ given $\gamma_0$, $\beta_{lj}$ and $A$:] 
The update expression for $\gamma_j$ is given by
\begin{eqnarray}
\hat{\gamma}_{j^{\prime}} \leftarrow \frac{S \left( (1-w) \sum_{i=1}^n y^{**}_i x^*_{ij^{\prime}}, \frac{\lambda_{\gamma}}{2} \right) }{(1-w) \sum_{i=1}^n x^{*2}_{i j^{\prime}}}, \quad (j^{\prime} = 1,\ldots, k),
\label{UpdateGamma}
\end{eqnarray}
where
\begin{eqnarray*}
x^*_{ij} &=& {\bm \beta}^T_j {\bm x}_i, \\
y^{**}_i &=& y_i - \gamma_0 - \sum_{j \neq j^{\prime}} \gamma_j x^*_{ij}.
\end{eqnarray*}

\item[ $A$ given $\gamma_0$, $\beta_{lj}$ and $\gamma_j$:]
The estimate of $A$ is obtained by 
\begin{eqnarray*}
\hat{A} = U V^T,
\end{eqnarray*}
where $(X^T X) B = U D V^T$.

\item[ $\gamma_0$ given $\beta_{lj}$, $\gamma_j$ and $A$:]
The estimate of $\gamma_0$ is derived from 
\begin{eqnarray*}
\hat{\gamma}_0 = \frac{1}{n} \sum_{i=1}^n \left\{ y_i - \sum_{j=1}^k \hat{\gamma}_j \left( \sum_{l=1}^p \hat{\beta}_{lj} x_{il} \right) \right\}.
\end{eqnarray*}
\end{description}
These procedures are iterated until convergence. 

\subsection{More efficient algorithm}
To speed up our algorithm, we apply the covariance updates, which was proposed by Friedman \textit{et al.} (2010), into the parameter updates.

We can rewrite the update of the parameter $B$ in (\ref{UpdateBeta}) in the form
\begin{eqnarray*}
 \sum_{i=1}^n x_{i l^{\prime}} \left\{ (1-w) Y_i \gamma_{j^{\prime}} + Y_{j^{\prime} i }^* w\right\} &=& (1-w) \gamma_{j^{\prime}} \sum_{i=1}^n x_{il^\prime} r_i + w \sum_{i=1}^n x_{i l^\prime} r^*_{j^\prime i} \\
&& \hspace{15mm} + \tilde{\beta}_{l^\prime j^\prime} \sum_{i=1}^n x_{i l^\prime}^2 \left\{ (1-w) \gamma_{j^\prime}^2 + w \right\}, 
\end{eqnarray*}
where $\tilde{\beta}_{l^\prime j^\prime}$ is the current estimate of ${\beta}_{l^\prime j^\prime}$, $r_i = y_i - \gamma_0 - \sum_{j=1}^k \sum_{l=1}^p \gamma_j \tilde{\beta}_{lj} x_{il}$ and $r^*_{j^\prime i} = y^*_{j^\prime i} - \sum_{l=1}^p \tilde{\beta}_{l j^\prime} x_{i l}$. 
After simple calculation, the first term on the right-hand side (up to $(1-w) \gamma_{j^{\prime}} $) becomes
\begin{eqnarray}
\sum_{i=1}^n x_{il^\prime} r_i = \sum_{i=1}^n x_{i l^\prime} y_i - \gamma_0 \sum_{i=1}^n x_{i l^\prime} - \sum_{j,l : |\tilde{\beta}_{lj}| > 0} \gamma_j \tilde{\beta}_{l j} {\bm x}_l^T {\bm x}_l,
\label{CovUp1}
\end{eqnarray}
and the second term on the right-hand side (up to $w$) is
\begin{eqnarray}
\sum_{i=1}^n x_{i l^\prime} r^*_{j^\prime i} = \sum_{i=1}^n x_{i l^\prime} y^*_{j^\prime i} - \sum_{l : |\tilde{\beta}_{l j^\prime}|>0} \tilde{\beta}_{l j^\prime} {\bm x}_l^T {\bm x}_l.
\label{CovUp2}
\end{eqnarray}
These formulas largely reduces computational task, because we update only the last term on (\ref{CovUp1}) and (\ref{CovUp2}) when the estimate of $\beta_{l^\prime j^\prime}$ is non-zero, while we do not update (\ref{CovUp1}) and (\ref{CovUp2}) when the estimate of $\beta_{l^\prime j^\prime}$ is zero. 

Similarly, the update of the parameter ${\bm \gamma}$ in  (\ref{UpdateGamma}) is written as
\begin{eqnarray}
\sum_{i=1}^n y_i^{**} x_{i j^\prime}^* = \sum_{i=1}^n s_i x_{i j^\prime}^* + \tilde{\gamma}_{j^\prime} \sum_{i=1}^n x^{*2}_{i j^\prime},
\end{eqnarray}
where $\tilde{\gamma}_{j^\prime}$ is the current estimate of ${\gamma}_{j^\prime}$. 
The first term on the right becomes
\begin{eqnarray}
\sum_{i=1}^n s_i x_{i j^\prime}^* = \sum_{i=1}^n y_i x_{i j^\prime}^* - \gamma_0 \sum_{i=1}^n x_{i j^\prime}^* - \sum_{j : |\tilde{\gamma}_j| > 0} \tilde{\gamma}_j {\bm x}_j^{*T} {\bm x}_j^{*}. 
\label{CovUp3}
\end{eqnarray}
Therefore we update only the last term on (\ref{CovUp3}) when the estimate of $\gamma_{j^\prime}$ is non-zero, while we do not update (\ref{CovUp3}) when the estimate of $\gamma_{j^\prime}$ is zero. 

\subsection{Selection of tuning parameters}

SPCR and aSPCR depend on four tuning parameters $(w, \zeta, \lambda_{\beta}, \lambda_{\gamma})$. 
To avoid this hard computational task, we fix the values of $w$ and $\zeta$, and then optimize only two tuning parameters $\lambda_{\beta}$ and $\lambda_{\gamma}$.

The tuning parameter $w$ plays a role in prediction accuracy. 
While a smaller value for $w$ provides good prediction, the estimated models often tend to be unstable due to the flexibility of $B$. 
We tried many simulations with several values for $w$, and then we concluded to set $w=0.1$ in this study. 
The tuning parameter $\zeta$ takes the role in the trade-off between the $L_1$ and $L_2$ penalties on $B$. 
The value of $\zeta$ in elastic net is usually determined by users (Zou and Hastie, 2005). 
In our simulation studies we fixed $\zeta$ as 0.01.  

The tuning parameters $\lambda_{\beta}$ and $\lambda_{\gamma}$ are optimized using $K$-fold cross-validation. 
When the original dataset is divided into $K$ datasets  $({\bm y}^{(1)}, X^{(1)}), \ldots, ({\bm y}^{(K)}, X^{(K)})$, the CV criterion is given by
\begin{eqnarray*}
{\rm CV} = \frac{1}{K} \sum_{k=1}^K || {\bm y}^{(k)} - \hat{\gamma}_0^{(-k)} {\bm 1}_{(k)} - X^{(k)} \hat{B}^{(-k)} \hat{\bm \gamma}^{(-k)} ||^2,
\end{eqnarray*}
where ${\bm 1}_{(k)}$ is a vector of which the elements are all one, and $\hat{\gamma}_0^{(-k)}, \hat{B}^{(-k)}, \hat{\bm \gamma}^{(-k)}$ are the estimates computed with the data removing the $k$-th part. 
We employed $K=5$ in our simulation. 
The tuning parameters $\lambda_{\beta}$ and $\lambda_{\gamma}$ were, respectively, selected from 10 equally-spaced values on $[ \lambda_{\min}, \lambda_{\max} ]$, where $\lambda_{\min}$ and $\lambda_{\max}$ were determined  according to the function \texttt{glmnet} in \texttt{R}.


\section{Numerical study}
\label{NumericalStudy}

\subsection{Monte Carlo simulations}
\label{MonteCarlo}

Monte Carlo simulations were conducted to investigate the performances of our proposed method. 
Three models were examined in this study. 

In the first model, we considered the 10-dimensional covariate vector ${\bm x} =(x_1,\ldots,x_{10})^T$ according to a multivariate normal distribution with mean zero vector and variance-covariance matrix $\Sigma_1$, and  generated the response $y$ from the linear regression model given by
\begin{eqnarray*}
y_i = \xi^*_1 { x}_{i1} + \xi_2^* {x}_{i2} + \varepsilon_i, \quad \varepsilon_i \sim N(0, \sigma^2), \quad i=1,\ldots,n. 
\end{eqnarray*}
\textcolor{black}{
We used 
$\xi_1^* = 2, \xi_2^* = 1, \Sigma_1 =I_{10}$ (Case 1(a)), where $I_{10}$ is  the $10 \times 10$ identity matrix, and 
$\xi_1^* = 8, \xi_2^* = 1, \Sigma_1 ={\rm diag} (1,3^2,\ldots,1)$ (Case 1(b)).
Case 1(a) is a simple situation. Case 1(b) corresponds to the situation discussed in Sect. \ref{Related work}.
}

In the second model, we considered the 20-dimensional covariate vector ${\bm x} =(x_1,\ldots,x_{20})^T$ according to a multivariate normal distribution $N({\bm 0}_{20}, \Sigma_2)$, and generated the response $y$ \textcolor{black}{by}
\begin{eqnarray*}
y_i = 4 {\bm x}_i^T {\bm \xi}^* + \varepsilon_i, \quad \varepsilon_i \sim N(0,\sigma^2), \quad i=1,\ldots,n.
\end{eqnarray*}
\textcolor{black}{
We used $\Sigma_2 = {\rm block diag} (\Sigma_{2}^*, I_{11})$ and ${\bm \xi}^* = ({\bm \nu}_1^*,0,\ldots,0)^T$, where $\left( \Sigma_{2}^* \right)_{ij} = 0.9^{|i-j|} \ (i,j=1,\ldots,9)$ and  ${\bm \nu}_1^*=(-1,0,1,1,0,-1,-1,0,1)$ is a sparse approximation of the fourth eigenvector of $\Sigma_{2}^*$ (Case 2). 
This case deals with the situation where the response is associated with the principal component loading with small eigenvalue. Note that even if each explanatory variable ${\bm x}$ is normalized, the principal component $\bm{x}^T \bm{\xi}$ does not have unit variance in general.
}


In the third model, we assumed the 30-dimensional covariate vector ${\bm x} =(x_1,\ldots,x_{30})^T$ according to a multivariate normal distribution $N({\bm 0}_{30}, \Sigma_3)$, and generated the response $y$ \textcolor{black}{by}
\begin{eqnarray*}
y_i = 4 {\bm x}_i^T {\bm \xi}^{*}_1 +  4{\bm x}_i^T {\bm \xi}^{*}_2 + \varepsilon_i, \quad \varepsilon_i \sim N(0,\sigma^2), \quad i=1,\ldots,n.
\end{eqnarray*}
We used $\Sigma_3={\rm block diag} (\Sigma_{2}^*, \Sigma_{3}^*, I_{15})$ with $\left( \Sigma_{3}^* \right)_{ij} = 0.9^{|i-j|} \ (i,j=1,\ldots,6)$, \textcolor{black}{and ${\bm \xi}^{*}_1 = ({\bm \nu}^{*}_{1}, 0,\ldots,0)^T$}. 
Two cases were considered for 
\textcolor{black}{${\bm \xi}_2^*=(0,\ldots,0,{\bm \nu}_{2}^*,0,\ldots,0)^T$}, where the first nine and last 15 values are zero. 
First, we used \textcolor{black}{${\bm \nu}^*_{2} = (\underbrace{1, \ldots,1}_{6})$} that is an approximation of the first eigenvector of $\Sigma_{3}^*$ (Case 3(a)). 
Second, we used \textcolor{black}{${\bm \nu}^*_{2} = (1, 0, -1, -1, 0, 1)$} that is a sparse approximation of the third eigenvector of $\Sigma_{3}^*$ (Case 3(b)).
\textcolor{black}{
Case 3 is a more complex situation.
}

The sample size was set to $n=50, 200$. 
The standard error $\sigma$ was set to 0.1 or 1. 
Our proposed methods, SPCR and aSPCR, were fitted to the simulated data with one or 10 components $(k=1,10)$ for Case 1, one or five components $(k=1,5)$ for Case 2, and 10 components $(k=10)$ for Case 3. 
Our proposed methods were compared with SPLS, PLS, and PCR. 
SPLS was computed by the package {\ttfamily spls} in {\ttfamily R}, and PLS and PCR by the package {\ttfamily pls} in {\ttfamily R}.
The number of components and the values of tuning parameters in SPLS, PLS, and PCR were selected by 10-fold cross-validation. 
The performance was evaluated by ${\rm MSE} = E \left[ (y - \hat{y})^2 \right]$. 
The simulation was conducted 100 times and the MSE was estimated by 1,000 random samples.

\begin{table}[htbp]
\begin{center}
\small
\caption{Mean (standard deviation) of MSE for $\sigma=0.1$.
The bold values correspond to the smallest mean. }
\vspace{5mm}
\begin{tabular}{@{\extracolsep{-5.5pt}}lcccccccc} \hline
Case & $k$ & $n$ & aSPCR & SPCR & SPLS & PLS & PCR \\ \hline
1(a) & 1 & 50 & ${\bf 1.095\times 10^{-2}}$ & $1.654\times 10^{-1}$ & $2.952\times 10^{-1}$ & $8.877\times 10^{-1}$ &  4.643 \\
 & & & $(9.906\times 10^{-4})$ & ($8.799\times 10^{-1}$) & $(3.919\times 10^{-1})$ & $(3.885\times 10^{-1})$ &  $(6.325\times 10^{-1})$ \\
 & & 200 & ${\bf 1.019\times 10^{-2}}$ & $5.735\times 10^{-2}$ & $3.167\times 10^{-2}$ & $2.249\times 10^{-1}$ &  4.605 \\
 & & & $(5.088\times 10^{-4})$ & ($4.702\times 10^{-1}$) & $(3.095\times 10^{-2})$ & $(9.559\times 10^{-2})$ & $(5.240\times 10^{-1})$ \\
 & 10 & 50 & $1.156\times 10^{-2}$ & $1.162\times 10^{-2}$ & ${\bf 1.118\times 10^{-2}}$ & $1.283\times 10^{-2}$ & $1.282\times 10^{-2}$ \\
 & & & $(1.072\times 10^{-3})$ & $(1.107\times 10^{-3})$ & $(1.304\times 10^{-3})$ & $(1.380\times 10^{-3})$ & $(1.379\times 10^{-3})$ \\
 & & 200 & $1.029\times 10^{-2}$ & $1.031\times 10^{-2}$ & ${\bf 1.021\times 10^{-2}}$ &  $1.054\times 10^{-2}$ & $1.054\times 10^{-2}$ \\
 & & & $(5.063\times 10^{-4})$ & $(5.628\times 10^{-4})$ & $(5.120\times 10^{-4})$ & $(5.216\times 10^{-4})$ &  $(5.218\times 10^{-4})$ \\ \hline
1(b) & 1 & 50 &${\bf 1.250\times 10^{-2}}$ & $1.465\times 10^{-2}$ & $4.043\times 10^{1}$ & $4.595\times 10^{1}$ & $6.650\times 10^{1}$ \\
 & & & $(2.220\times 10^{-3})$ & $(2.778 \times 10^{-3})$ & $(1.869\times 10^{1})$ & $(1.148\times 10^{1})$ &  $(4.517)$ \\
 & & 200 &${\bf 1.131\times 10^{-2}}$ & $1.186\times 10^{-2}$ & $3.975\times 10^{1}$ & $4.532\times 10^{1}$ &  $6.457\times 10^{1}$ \\
 & & & $(7.155\times 10^{-4})$ & $(7.808\times 10^{-4})$ & $(1.531\times 10^{1})$ & $(5.048)$ &  $(2.919)$ \\
 & 10 & 50 & $1.140\times 10^{-2}$ & $1.156\times 10^{-2}$ & ${\bf 1.126\times 10^{-2}}$ & $1.284\times 10^{-2}$ & $1.282\times 10^{-2}$ \\
 & & & $(1.132\times 10^{-3})$ & $(1.222\times 10^{-3})$ & $(1.508\times 10^{-3})$ & $(1.395\times 10^{-3})$ & $(1.379\times 10^{-3})$ \\
 & & 200 & $1.029\times 10^{-2}$ & $1.026\times 10^{-2}$ & ${\bf 1.023\times 10^{-2}}$ & $1.054\times 10^{-2}$ & $1.054\times 10^{-2}$ \\
 & & & $(5.258\times 10^{-4})$ & $(5.526\times 10^{-4})$ & $(4.955\times 10^{-4})$ & $(5.223\times 10^{-4})$ & $(5.218\times 10^{-4})$ \\ \hline
2 & 1 & 50 & ${\bf 1.241\times 10^{-2}}$ & $1.614\times 10^{-2}$ & $1.978\times 10^{1}$ & $1.979\times 10^{1}$ &  $2.038\times 10^{1}$ \\
 & & & $(1.738\times 10^{-3})$ & $(3.601\times 10^{-3})$ & $(1.909)$ & $(1.851)$ & $(1.272)$ \\
 & & 200 & ${\bf 1.051\times 10^{-2}}$ & $1.102\times 10^{-2}$ & $1.418\times 10^{1}$ & $1.571\times 10^{1}$ & $1.967\times 10^{1}$ \\
 & & & $(6.754\times 10^{-4})$ & ($8.276\times 10^{-4}$) & $(4.475)$ & $(2.938)$ & $(8.374\times 10^{-1})$ \\
 & 5 & 50 & ${\bf 1.313\times 10^{-2}}$ & $1.548\times 10^{-2}$ & $3.946\times 10^{-1}$ & 1.946 & $2.118\times 10^{1}$ \\
 & & & $(2.207\times 10^{-3})$ & $(3.708\times 10^{-3})$ & ($6.452\times 10^{-1}$) & $(1.337)$ & $(1.426)$ \\
 & & 200 & ${\bf 1.077\times 10^{-2}}$ & $1.091\times 10^{-2}$ & $1.667\times 10^{-2}$ & $8.268\times 10^{-2}$ & $1.978\times 10^{1}$ \\
 & & & $(7.140\times 10^{-4})$ & ($7.768\times 10^{-4}$) & $(1.274\times 10^{-2})$ & $(4.039\times 10^{-2})$ & $(8.926\times 10^{-1})$ \\ \hline
3(a) & 10 & 50 & ${\bf 1.831\times 10^{-2}}$ & $2.191\times 10^{-2}$ & $3.438\times 10^{-1}$ & $8.493\times 10^{-1}$ & $2.839\times 10^{1}$ \\
 & & & $(4.842\times 10^{-3})$ & $(6.641\times 10^{-3})$ & $(4.319\times 10^{-1})$ & $(6.014\times 10^{-1})$ & $(5.090)$ \\
 & & 200 & ${\bf 1.158\times 10^{-2}}$ & $1.166\times 10^{-2}$ & $1.247\times 10^{-2}$ & $2.407\times 10^{-2}$ & $2.172\times 10^{1}$  \\
 & & & $(8.208\times 10^{-4})$ & $(8.225\times 10^{-4})$ & $(1.597\times 10^{-3})$ & $(7.115\times 10^{-3})$ & ($1.463\times 10^{-1}$) \\ \hline
3(b) & 10 & 50 & ${\bf 1.721\times 10^{-2}}$ & $2.180\times 10^{-2}$ & $4.852\times 10^{-1}$ & 1.295 & $3.676\times 10^{1}$ \\
 & & & $(5.311\times 10^{-3})$ & $(6.390\times 10^{-3}$) & $(6.966\times 10^{-1})$ & $(9.401\times 10^{-1})$ & $(2.676)$ \\
 & & 200 & $1.185\times 10^{-2}$ & ${\bf 1.167\times 10^{-2}}$ & $1.201\times 10^{-2}$ & $2.972\times 10^{-2}$ & $3.373\times 10^{1}$ \\
 & & & $(9.778\times 10^{-4})$ & $(8.533\times 10^{-4})$ & $(1.710\times 10^{-3})$ & $(1.030\times 10^{-2})$ & $(1.605)$ \\ \hline
\end{tabular}
\label{Table_MSE_sigma01}
\end{center}
\end{table}

\begin{table}[htbp]
\begin{center}
\small
\caption{Mean (standard deviation) of MSE for $\sigma=1$.
The bold values correspond to the smallest mean. }
\vspace{5mm}
\begin{tabular}{@{\extracolsep{-5.5pt}}lcccccccc} \hline
Case & $k$ & $n$ & aSPCR & SPCR & SPLS & PLS & PCR \\ \hline
1(a) & 1 & 50 & {\bf 1.266} & 1.638 & 1.475 & 1.999 &  5.663 \\
 & & & $(8.134\times 10^{-1})$ & $(1.361)$ & $(4.789\times 10^{-1})$ & $(4.331\times 10^{-1})$ &  $(6.464\times 10^{-1})$ \\
 & & 200 & 1.159 & 1.333 & {\bf 1.031} & 1.256 &  5.598 \\
 & & & $(8.267\times 10^{-1})$ & $(1.169)$ & $(5.665\times 10^{-2})$ & $(1.225\times 10^{-1})$ & $(5.593\times 10^{-1})$ \\
 & 10 & 50 & 1.123 & 1.194 & {\bf 1.122} & 1.283 & 1.282 \\
 & & & $(1.163\times 10^{-1})$ & $(1.142\times 10^{-1})$ & $(1.357\times 10^{-1})$ & $(1.388\times 10^{-1})$ & $(1.377\times 10^{-2})$ \\
 & & 200 & 1.023 & 1.034 & {\bf 1.021} &  1.054 & 1.054 \\
 & & & $(4.983\times 10^{-2})$ & $(5.214\times 10^{-2})$ & $(5.136\times 10^{-2})$ & $(5.208\times 10^{-2})$ &  $(5.218\times 10^{-2})$ \\ \hline
1(b) & 1 & 50 &{\bf 1.191} & 1.283 & $4.144\times 10^{1}$ & $4.711\times 10^{1}$ & $6.748\times 10^{1}$ \\
 & & & $(1.260\times 10^{-1})$ & $(1.383 \times 10^{-1})$ & $(1.871\times 10^{1})$ & $(1.137\times 10^{1})$ &  $(4.646)$ \\
 & & 200 &{\bf 1.030} & 1.062 & $4.050\times 10^{1}$ & $4.629\times 10^{1}$ &  $6.560\times 10^{1}$ \\
 & & & $(5.226\times 10^{-2})$ & $(5.493\times 10^{-2})$ & $(1.565\times 10^{1})$ & $(5.246)$ &  $(3.078)$ \\
 & 10 & 50 & {\bf 1.139} & 1.194 & 1.149 & 1.315 & 1.314 \\
 & & & $(1.450\times 10^{-1})$ & $(1.569\times 10^{-1})$ & $(1.626\times 10^{-1})$ & $(1.662\times 10^{-1})$ & $(1.658\times 10^{-1})$ \\
 & & 200 & {\bf 1.023} & 1.035 & {\bf 1.023} & 1.054 & 1.054 \\
 & & & $(5.204\times 10^{-2})$ & $(5.573\times 10^{-2})$ & $(5.238\times 10^{-2})$ & $(5.221\times 10^{-2})$ & $(5.218\times 10^{-2})$ \\ \hline
2 & 1 & 50 & {\bf 1.284} & 1.583 & $2.079\times 10^{1}$ & $2.084\times 10^{1}$ &  $2.140\times 10^{1}$ \\
 & & & $(2.522\times 10^{-1})$ & $(3.245\times 10^{-1})$ & $(1.788)$ & $(2.012)$ & $(1.295)$ \\
 & & 200 & {\bf 1.058} & 1.120 & $1.568\times 10^{1}$ & $1.695\times 10^{1}$ & $2.086\times 10^{1}$ \\
 & & & $(5.566\times 10^{-2})$ & $(6.347\times 10^{-2})$ & $(4.475)$ & $(2.981)$ & $(8.458\times 10^{-1})$ \\
 & 5 & 50 & {\bf 1.279} & 1.576 & 2.017 & 3.398 & $2.224\times 10^{1}$ \\
 & & & $(2.434\times 10^{-1})$ & $(3.221\times 10^{-1})$ & $(1.048)$ & $(1.442)$ & $(1.476)$ \\
 & & 200 & {\bf 1.060} & 1.119 & 1.075 & 1.175 & $2.097\times 10^{1}$ \\
 & & & $(5.671\times 10^{-2})$ & $(6.323\times 10^{-2})$ & $(5.837\times 10^{-2})$ & $(7.427\times 10^{-2})$ & $(8.876\times 10^{-1})$ \\ \hline
3(a) & 10 & 50 & {\bf 1.607} & 2.274 & 2.403 & 2.724 & $2.961\times 10^{1}$ \\
 & & & $(4.250\times 10^{-1})$ & $(6.044\times 10^{-1})$ & $(8.958\times 10^{-1})$ & $(7.205\times 10^{-1})$ & $(5.070)$ \\
 & & 200 & {\bf 1.088} & 1.162 & 1.156 & 1.187 & $2.277\times 10^{1}$  \\
 & & & $(7.104\times 10^{-2})$ & $(7.882\times 10^{-2})$ & $(2.621\times 10^{-1})$ & $(7.714\times 10^{-2})$ & $(1.539)$ \\ \hline
3(b) & 10 & 50 & {\bf 1.482} & 2.180 & 2.364 & 3.081 & $3.793\times 10^{1}$ \\
 & & & $(3.094\times 10^{-1})$ & $(5.990\times 10^{-1})$ & $(9.068\times 10^{-1})$ & $(8.959\times 10^{-1})$ & $(2.835)$ \\
 & & 200 & {\bf 1.085} & 1.165 & 1.158 & 1.192 & $3.482\times 10^{1}$ \\
 & & & $(6.686\times 10^{-2})$ & $(7.719\times 10^{-2})$ & $(4.742\times 10^{-1})$ & $(7.631\times 10^{-2})$ & $(1.698)$ \\ \hline
\end{tabular}
\label{Table_MSE_sigma1}
\end{center}
\end{table}

\renewcommand{\arraystretch}{0.87}

\begin{table}[htbp]
\begin{center}
\small
\caption{Mean (standard deviation) of TPR and TNR for $\sigma=0.1$.
The bold values correspond to the largest TPR and TNR. }
\vspace{5mm}
\begin{tabular}{@{\extracolsep{3pt}}lcccccccc} \hline
           &        &         &                &TPR     &             &          &TNR&\\
    \cline{4-6}  \cline{7-9}
Case & $k$ & $n$ & aSPCR & SPCR & SPLS & aSPCR & SPCR & SPLS \\ \hline
1(a) & 1 & 50 &{\bf 1} & 0.970 & 0.930 & {\bf 1} & 0.615 &0.982\\
 & & & $(0)$ & $(0.171)$ & $(0.174)$ & $(0)$ &  $(0.285)$ &$(0.053)$\\
 & & 200 &{\bf 1} & 0.990 & {\bf 1} & {\bf 1} & 0.631 &{\bf 1} \\
 & & & $(0)$ & $(0.100)$ & $(0)$ & $(0)$ &  $(0.318)$ &$(0)$\\
 & 10 & 50 & {\bf 1} & {\bf 1} & {\bf 1} & 0.693 & 0.496 &{\bf 0.930} \\
 & & & $(0)$ & $(0)$ & $(0)$ & $(0.368)$ &  $(0.287)$ &$(0.130)$\\
 & & 200 & {\bf 1} & {\bf 1} & {\bf 1} & 0.562 & 0.528 &{\bf 0.911} \\
 & & & $(0)$ & $(0)$ & $(0)$ & $(0.316)$ &  $(0.265)$ &$(0.160)$\\ \hline
1(b) & 1 & 50 & {\bf 1} & {\bf 1} & 0.870 & {\bf 1} & 0.061 &0.926\\
 & & & $(0)$ & $(0)$ & $(0.220)$ & $(0)$ &  $(0.158)$ &$(0.158)$\\
 & & 200 &{\bf 1} & {\bf 1} & 0.905 & {\bf 1} & 0.070 &0.963\\
 & & & $(0)$ & $(0)$ & $(0.197)$ & $(0)$ &  $(0.089)$ &$(0.087)$\\
 & 10 & 50 &{\bf 1} & {\bf 1} & {\bf 1} & 0.773 & 0.541 &{\bf 0.912}\\
 & & & $(0)$ & $(0)$ & $(0)$ & $(0.349)$ &  $(0.324)$ &$(0.195)$\\
 & & 200 &{\bf 1} & {\bf 1} & {\bf 1} & 0.698 & 0.688 &{\bf 0.897}\\
 & & & $(0)$ & $(0)$ & $(0)$ & $(0.329)$ &  $(0.341)$ &$(0.156)$\\ \hline
2 & 1 & 50 & {\bf 1} & {\bf 1} & 0.548 & {\bf 1} & 0.267 &0.718 \\
 & & & $(0)$ & $(0)$ & $(0.285)$ & $(0)$ &  $(0.172)$ &$(0.312)$\\
 & & 200 & {\bf 1} & {\bf 1} & 0.861 & {\bf 1} & 0.336 &0.817 \\
 & & & $(0)$ & $(0)$ & $(0.174)$ & $(0)$ &  $(0.166)$ &$(0.213)$\\
 & 5 & 50 & {\bf 1} & {\bf 1} & 0.995 & {\bf 0.859} & 0.304 &0.775 \\
 & & & $(0)$ & $(0)$ & $(0.028)$ & $(0.111)$ &  $(0.196)$ &$(0.135)$\\
 & & 200 & {\bf 1} & {\bf 1} & {\bf 1} & 0.905 & 0.387 &{\bf 0.931} \\
 & & & $(0)$ & $(0)$ & $(0)$ & $(0.075)$ &  $(0.252)$ &$(0.073)$\\ \hline
3(a) & 10 & 50 & {\bf 1} & {\bf 1} & {\bf 1} & {\bf 0.862} & 0.289 &0.503 \\
 & & & $(0)$ & $(0)$ & $(0)$ & $(0.102)$ &  $(0.168)$ &$(0.146)$\\
 & & 200 & {\bf 1} & {\bf 1} & {\bf 1} & {\bf 0.903} & 0.316 &0.816 \\
 & & & $(0)$ & $(0)$ & $(0)$ & $(0.062)$ &  $(0.216)$ &$(0.079)$\\ \hline
3(b) & 10 & 50 & {\bf 1} & {\bf 1} & 0.998 & {\bf 0.854} & 0.271 &0.516 \\
 & & & $(0)$ & $(0)$ & $(0.014)$ & $(0.092)$ &  $(0.155)$ &$(0.165)$\\
 & & 200 & {\bf 1} & {\bf 1} & {\bf 1} & {\bf 0.916} & 0.294 &0.822 \\
 & & & $(0)$ & $(0)$ & $(0)$ & $(0.061)$ &  $(0.182)$ &$(0.083)$\\ \hline
\end{tabular}
\label{Table_TPTN_sigma01}
\end{center}
\end{table}

\begin{table}[htbp]
\begin{center}
\small
\caption{Mean (standard deviation) of TPR and TNR for $\sigma=1$.
The bold values correspond to the largest TPR and TNR. }
\vspace{5mm}
\begin{tabular}{@{\extracolsep{3pt}}lcccccccc} \hline
           &        &         &                &TPR     &             &          &TNR&\\
    \cline{4-6}  \cline{7-9}
Case & $k$ & $n$ & aSPCR & SPCR & SPLS & aSPCR & SPCR & SPLS \\ \hline
1(a) & 1 & 50 &{\bf 0.970} & 0.910 & 0.910 & 0.791 & 0.258 &{\bf 0.953}\\
 & & & $(0.171)$ & $(0.287)$ & $(0.193)$ & $(0.247)$ &  $(0.277)$ &$(0.128)$\\
 & & 200 &0.970 & 0.940 & {\bf 1} & 0.870 & 0.250 &{\bf 0.998} \\
 & & & $(0.171)$ & $(0.238)$ & $(0)$ & $(0.183)$ &  $(0.255)$ &$(0.012)$\\
 & 10 & 50 & {\bf 1} & 0.990 & {\bf 1} & 0.802 & 0.227 &{\bf 0.931} \\
 & & & $(0)$ & $(0.100)$ & $(0)$ & $(0.334)$ &  $(0.168)$ &$(0.141)$\\
 & & 200 & {\bf 1} & {\bf 1} & {\bf 1} & 0.737 & 0.318 &{\bf 0.911} \\
 & & & $(0)$ & $(0)$ & $(0)$ & $(0.353)$ &  $(0.204)$ &$(0.164)$\\ \hline
1(b) & 1 & 50 &{\bf 1} & {\bf 1} & 0.870 & 0.550 & 0.012 &{\bf 0.915}\\
 & & & $(0)$ & $(0)$ & $(0.220)$ & $(0.219)$ &  $(0.057)$ &$(0.166)$\\
 & & 200 &{\bf 1} & {\bf 1} & 0.900 & 0.728 & 0.007 &{\bf 0.966}\\
 & & & $(0)$ & $(0)$ & $(0.201)$ & $(0.185)$ &  $(0.029)$ &$(0.083)$\\
 & 10 & 50 &{\bf 1} & {\bf 1} & {\bf 1} & 0.860 & 0.542 &{\bf 0.895}\\
 & & & $(0)$ & $(0)$ & $(0)$ & $(0.278)$ &  $(0.305)$ &$(0.187)$\\
 & & 200 &{\bf 1} & {\bf 1} & {\bf 1} & 0.831 & 0.525 &{\bf 0.900}\\
 & & & $(0)$ & $(0)$ & $(0)$ & $(0.366)$ &  $(0.349)$ &$(0.174)$\\ \hline
2 & 1 & 50 & {\bf 1} & {\bf 1} & 0.543 & {\bf 0.865} & 0.172 &0.726 \\
 & & & $(0)$ & $(0)$ & $(0.313)$ & $(0.182)$ &  $(0.139)$ &$(0.317)$\\
 & & 200 & {\bf 1} & {\bf 1} & 0.860 & {\bf 0.930} & 0.202 &0.775 \\
 & & & $(0)$ & $(0)$ & $(0.215)$ & $(0.122)$ &  $(0.153)$ &$(0.253)$\\
 & 5 & 50 & {\bf 1} & {\bf 1} & 0.993 & {\bf 0.872} & 0.176 &0.648 \\
 & & & $(0)$ & $(0)$ & $(0.032)$ & $(0.191)$ &  $(0.145)$ &$(0.200)$\\
 & & 200 & {\bf 1} & {\bf 1} & {\bf 1} & 0.892 & 0.205 &{\bf 0.896} \\
 & & & $(0)$ & $(0)$ & $(0)$ & $(0.190)$ &  $(0.150)$ &$(0.111)$\\ \hline
3(a) & 10 & 50 & 0.999 & {\bf 1} & 0.998 & {\bf 0.885} & 0.142 &0.423 \\
 & & & $(0.008)$ & $(0)$ & $(0.011)$ & $(0.148)$ &  $(0.101)$ &$(0.220)$\\
 & & 200 & {\bf 1} & {\bf 1} & 0.999 & {\bf 0.901} & 0.165 &0.846 \\
 & & & $(0)$ & $(0)$ & $(0.008)$ & $(0.164)$ &  $(0.122)$ &$(0.163)$\\ \hline
3(b) & 10 & 50 & {\bf 1} & {\bf 1} & 0.999 & {\bf 0.880} & 0.184 &0.430 \\
 & & & $(0)$ & $(0)$ & $(0.010)$ & $(0.130)$ &  $(0.128)$ &$(0.202)$\\
 & & 200 & {\bf 1} & {\bf 1} & 0.998 & {\bf 0.875} & 0.223 &0.864 \\
 & & & $(0)$ & $(0)$ & $(0.020)$ & $(0.203)$ &  $(0.162)$ &$(0.148)$\\ \hline
\end{tabular}
\label{Table_TPTN_sigma1}
\end{center}
\end{table}

\textcolor{black}{
Tables \ref{Table_MSE_sigma01} and \ref{Table_MSE_sigma1} show the means and standard deviations of MSEs for $\sigma=0.1,1$, and present similar results. 
PCR was clearly the worst. SPLS was better than PLS, and aSPCR was basically better than SPCR. 
Therefore, we compare our methods, SPCR and aSPCR, with SPLS in more details. 
}

\textcolor{black}{
In Case 1(a), aSPCR was basically better than SPLS for $k=1$ and competitive to SPLS for $k=10$. 
In Case 1(b), SPCR and aSPCR provided much smaller MSEs than SPLS for $k=1$ and were competitive to SPLS for $k=10$. 
The results for $k=1$ correspond to that discussed in Sect. \ref{Related work}.  SPCR and aSPCR could appropriately select the loading related to the response. 
}

\textcolor{black}{
In Case~2, SPCR and aSPCR provided much smaller MSEs than SPLS for $k=1$, like in Case~1(b) for $k=1$, and aSPCR was better than SPLS for $k=5$. 
In addition, SPCR and aSPCR provided almost the same MSEs for $k=1$ as those for $k=5$. 
This means that SPCR and aSPCR could adaptively select the principal component loading with small eigenvalue. 
In Case~3, SPCR and aSPCR were better than SPLS. In complex situations for $n=50$, aSPCR outperforms SPLS. 
}
We also compared our methods with lasso, adaptive lasso (aLasso), elastic net (EN), and adaptive elastic net (aEN) (see the supplementary material). 
Our methods were better than or competitive with them, like SPLS was better than or competitive with them (Chun and Kele\c{s}, 2010).

We also computed the true positive rate (TPR) and the true negative rate (TNR) for aSPCR, SPCR, and SPLS, which are defined by
\begin{eqnarray*}
\mathrm{TPR}&=&
\frac{1}{100}
\sum_{k=1}^{100}
\frac{\left|\left\{ 
j:\hat{\xi}^{(k)}_{j}\neq0~\wedge~\xi^{\ast}_{j}\neq 0
\right\}\right|}
{\left|\left\{
j:\xi^{\ast}_{j}\neq 0 
\right\}\right|}, \\
\mathrm{TNR}&=&
\frac{1}{100}
\sum_{k=1}^{100}
\frac{\left|\left\{
j:\hat{\xi}^{(k)}_{j}=0~\wedge~\xi^{\ast}_{j}=0
\right\}\right|}
{\left|\left\{
j:\xi^{\ast}_{j}= 0 
\right\}\right|}, 
\end{eqnarray*}
where $\hat{\xi}^{(k)}_{j}$ is the estimated $j$-th coefficient for the $k$-th simulation, and $|\{\ast\}|$ is the number of elements included in a set $\{\ast\}$. 
Tables \ref{Table_TPTN_sigma01} and \ref{Table_TPTN_sigma1} show the means and standard deviations of TPR and TNR, and present similar results. 
In all cases, most of TPRs are very high. 
For TNR, SPLS provides higher ratios for simple situations (Cases 1(a) and 1(b)), while aSPCR provides higher ratios for complex situations (Cases 2, 3(a), and 3(b)). 
In particular, in Cases 3(a) and 3(b) for $n=50$, TNRs of aSPCR are much higher than those of SPLS.

\subsection{Real data analyses}
\label{RealData}

\begin{table}[t]
\begin{center}
\caption{Sample size and the numbers of covariates in real datasets.}
\vspace{5mm}
\begin{tabular}{lccccc}
\hline
          & housing  & energy & forest & concrete \\ \hline
sample size & 506   & 768 & 517 &  1030     \\
\# of covariates   & 13 & 8 &  10 &  8  \\
\hline
\end{tabular}
\label{Datasets}
\end{center}
\end{table}

We examined the effectiveness of our proposed method through real data analyses. 
Four benchmark datasets were used --- housing, energy, forest, and concrete. 
The datasets were obtained from the UCI database ({\it http://archive.ics.uci.edu/ml/index.html}). 
The sample size and  the numbers of covariates in these datasets are summarized in Table \ref{Datasets}. 
For the `energy' dataset, we used two types of response variables, following the explanation of the webpage. 
They are called `energy1' and `energy2' in this section.

First, using the `housing' dataset, we illustrate a behavior of aSPCR. 
The estimates $\hat{B}$ and $\hat{\bm \gamma}$ for aSPCR with $k=5$ were given by
\begin{eqnarray*}
&& \hat{\bm \beta}_1 = (0.025, 0.03, 0, 0, 0.058, 0, 0, 0, 0, 0.053, 0, 0, 0)^T, \quad \hat{\gamma}_1 = -33.87, \\
&& \hat{\bm \beta}_2 = (0, 0, 0, -0.007, 0, -0.032, 0, 0.036, -0.028, 0, 0.024, -0.009, 0.045)^T, \quad \hat{\gamma}_2 = -83.30.
\end{eqnarray*}
We also observed that $\hat{\gamma}_3 = \hat{\gamma}_4 = \hat{\gamma}_5 = 0$, and then $\hat{\bm \beta}_3$, $\hat{\bm \beta}_4$, and $\hat{\bm \beta}_5$ are omitted. 
The $L_1$-penalties on $B$ and ${\bm \gamma}$ could produce zero estimates for the parameters, and then caused automatic selection of principal components. 
In addition, we have
$$
\hat{B} \hat{\bm \gamma} =  (-0.87, 1.01, 0, 0.65, -1.97, 2.68, 0, -3.06, 2.38, -1.80, -2.03, 0.79, -3.75)^T,
$$
which suggests that the third and seven variables (that is, variables {\bf indus} and {\bf age}) are irrelevant with the response variable.  
This fact was pointed out in some literatures (see, e.g., Shao and Rao, 2000; Khalili, 2010; Leng, 2010).

\begin{table}[t]
\begin{center}
\small
\caption{Mean (standard deviation)  of MSE for real datasets. 
The bold values correspond to the smallest mean. }
\vspace{3mm}
\begin{tabular}{@{\extracolsep{-3pt}}lccccccccc} \hline
 & SPCR & aSPCR & SPLS & PLS & PCR & Lasso & aLasso & EN & aEN\\ \hline
housing & {\bf 28.94}	& 29.18 & 30.24& 29.78& 30.45	& 29.80& 30.16& 29.56& 30.05\\ 
             & (4.402) & (4.628) & (3.845)  & (4.467) &  (3.478)  & (4.055) &  (4.601)  & (3.953) &  (4.057) \\
energy1 & 11.07& 11.04& 11.14& 11.18& 14.90& 11.21& 11.06	& 11.18& {\bf 11.03}\\ 
             & (0.612) & (0.568) & (0.634)  & (0.627) &  (0.641)  & (0.637) &  (0.549)  & (0.624) &  (0.552) \\
energy2 & {\bf 9.248}& 9.275& 9.405& 9.386& 12.50& 9.379& 9.313& 9.340& 9.286\\ 
             & (0.468) & (0.519) & (0.532)  & (0.567) &  (0.523)  & (0.529) &  (0.521)  & (0.516) &  (0.527) \\
forest & 4680& 4569& 4579& 4683& 4599& {\bf 4534}& 4542& {\bf 4534}	& 4572\\ 
             & (466.6) & (757.4) & (652.0)  & (471.0) &  (600.7)  & (752.4) &  (709.6)  & (749.5) &  (628.5) \\
concrete & {\bf 121.2}& 121.3& 125.5& 122.0& 159.4& 122.1& 122.7& 122.7& 123.1\\  
             & (12.13) & (12.25) & (11.74)  & (11.67) &  (27.04)  & (12.08) &  (13.96)  & (12.36) &  (14.32) \\ \hline
\end{tabular}
\label{ResultRealDataAnal}
\end{center}
\end{table}

Next, for each dataset, we randomly used 100 observations as training data to estimate the parameters and the remaining observations as test data to estimate the MSE. 
The procedure was repeated 50 times. 
Our proposed method, SPCR and aSPCR, were compared with seven competing methods used in Sect. \ref{MonteCarlo}. 
The number of principal components or PLS components was set to  $k=5$ for aSPCR, SPCR, SPLS, PLS, and PCR. 
The tuning parameters $\lambda_{\beta}, \lambda_{\gamma}, \zeta$ in SPCR and aSPCR were selected by five-fold cross-validation; $\lambda_\beta$ and  $\lambda_\gamma$ were selected in similar manners to Sect. \ref{MonteCarlo} and $\zeta$ was selected from 0.1, 0.3, 0.5, 0.7, and 0.9. 
The tuning parameters in other methods were selected in similar manners to in Sect. \ref{MonteCarlo}.

Table \ref{ResultRealDataAnal} shows the means and standard deviations of MSEs. 
PCR was clearly worst except for the `forest' dataset. 
SPCR and aSPCR were better than other methods for the `housing', `energy2' and `concrete' datasets, and better or close to other methods for the `energy1' dataset. 
The MSEs for the `forest' dataset showed a different behavior from those for other datasets. 
PCR presented a good MSE, although aSPCR provided a smaller MSE than PCR. 
Furthermore, for all datasets, aSPCR was superior to SPLS, PLS, and PCR.


\section{Concluding remarks}
\label{Concludingremarks}

We proposed a one-stage procedure for PCR, which is constructed by combining a regression loss with PCA loss along with $L_1$ type regularization. 
We called this procedure SPCR. 
SPCR enabled us to adaptively provide sparse principal component loadings that are associated with the response and to select the number of principal components automatically. 
The estimation algorithm for SPCR was established via the coordinate decent algorithm. 
To obtain a more sparse regression model, we also proposed aSPCR, which assigns different weights to different parameters in the loading matrix $B$ in the estimation procedure. 
In numerical study, SPCR and aSPCR showed a good behavior in terms of prediction accuracy, TPR, and TNR. 
\section*{Acknowledgement}
This work was supported by the Bio-diversity Research Project of the Transdisciplinary Research Integration Center, Research Organization of Information and Systems. 



\end{document}